%% file: 2018_arxiv_psf.tex
\documentclass[oneside,english,reqno]{amsart}
\usepackage{amsmath}
\usepackage{amssymb}
\usepackage{euscript}
\usepackage{exscale,relsize}
\usepackage{graphicx}
\usepackage{srcltx}
\usepackage{xcolor}
\usepackage{subfigure}
\usepackage{multirow}
\usepackage{float}
\usepackage{algorithmic, algorithm}
 \newcommand{\RR}{\ensuremath{\mathbb{R}}}

 \newcommand{\prox}{\ensuremath{\operatorname{prox}}}

\newcommand{\argmin}{\operatorname{argmin}}

\newcommand{\regTermHyb}{\ensuremath{\psi}}
\newcommand{\dataTerm}{\ensuremath{\Phi}}
\newcommand{\operator}{H}
\newcommand{\inputSig}{x}

\newcommand{\outputSig}{y}

\newcommand{\Gnoise}{w}

\newcommand{\inputRV}{X}
\newcommand{\outputRV}{Y}
\newcommand{\GnoiseRV}{W}

\newcommand{\decomp}{V}

\newcommand{\psfPoint}{c}
\newcommand{\psfPSF}{P}

\newcommand{\minimize}[2]{\ensuremath{\underset{\substack{{#1}}}%
{\mathrm{minimize}}\;\;#2 }}
\newtheorem{problem}{Problem}
\newtheorem{model}{Model}

\begin{document}

\title{Spatially variant PSF modeling in confocal macroscopy}

\author{Anna Jezierska$^{1,2,3}$ \quad Hugues Talbot$^3$ \quad Jean-Christophe Pesquet$^4$ \quad Gilbert Engler$^5$}

\thanks{$^1$ Systems Research Institute Polish Academy of Sciences, 01-447 Warsaw, Poland \\
$^2$ Gdansk University of Technology, ETI, 80-233 Gdansk, Poland\\
$^3$ Universit{{\'e}} Paris-Est, LIGM, UMR CNRS 8049 - 77454 Marne-la-Vall{\'e}e, France \\
$^4$ Centrale Supélec, Centre pour la Vision Numérique, 92295 Chatenay-Malabry, France \\
$^5$ IBSV Unit, INRA - 06903 Sophia Antipolis, France\\
}
\maketitle
\tableofcontents{}
\begin{abstract}
Point spread function (PSF) plays an essential role in image reconstruction. 
In the context of confocal microscopy, optical performance
degrades towards the edge of the field of view as astigmatism, coma and vignetting. Thus, one should expect the related artifacts to be even stronger in macroscopy, where the field of view is much larger. The field aberrations in macroscopy fluorescence imaging system was observed to be symmetrical and to increase with the distance from the center of the field of view.  In this paper we propose an experiment and an optimization method for assessing the center of the field of view.
The obtained results constitute a step towards reducing the number of parameters in macroscopy PSF model. 
\end{abstract}

\section{Introduction}

The PSF determination is a crucial preliminary step to image restoration \cite{McNally_1999_3DImsgingDeconvolution}.  
Even if one resorts to blind deconvolution schemes, a priori knowledge related to the PSF is desired~\cite{Thiebaut_2002_optimization_issues_in_blind_deconvolution}, \cite{Bolte_2010_blind_deconvolution}, \cite{Soulez_2012_blind_deconvolution_in_widefield}. This knowledge can be acquired by studying PSF theoretical properties. In the context of fluorescence imaging, the theoretical approach usually relies on diffraction-limited PSF model~\cite{Kirshner_2012_PSF_fitting}. 
Experimental PSFs may be measured using calibration beads \cite{Yoo_2006_measurement_PSF} or directly from the image by extracting small point-like objects \cite{Tiedemann_2006_adaptive_PSF_estimation}. Such PSFs can be used for instance to validate theoretical parametric PSF model or to assess the aberration of point spread function in given imaging systems \cite{Pankajakshan_2012_macroscopy_PSF_journal}. 
The PSF modeling problem becomes more complex if the PSF is not spatially invariant.
The space variation model usually relies on one of the following strategies. Firstly, assuming that the PSF variation is smooth, the PSF can be represented as a weighted sum of basis functions~\cite{Arigovindan_2010_widefield_PSF_modeling}.
The efficiency can be further improved by applying interpolation methods~\cite{Denis_2011_PSF_modelling}. Alternatively an image 
can be segmented into regions inside which PSFs are assumed to be invariant~\cite{Rerabek_2008_space_variant_PSF}. Recently in \cite{Thiebaut_2016_psf_spatiall_variant} the authors shown experimentally that the first strategy leads to the better results. While in this work the focus was on astronomical images, we will concentrate on macroscopy.

 In high angular resolution images the PSF
 varies in the field of view, i.e. the optical aberrations increase towards the margins. This phenomena occurs in the context of astronomy~\cite{Denis_2011_PSF_modelling}, \cite{Thiebaut_2016_psf_spatiall_variant} (2D-PSF) or macroscopy \cite{Jezierska_phd}, where the principal axis of the PSFs around each bead were observed to converge to one point, called here the optical center.
 In macroscopy the problem of field aberration is coupled
with the problem of out-of-focus blur, in the depth-direction, due to the
diffraction-limited nature of the lens.
3D-PSF model for confocal macroscopy was previously studied in \cite{Panjakshan_2010_macroscope_PSF_modelling_Asilomar}.
One limitation of the proposed PSF model is that it requires two parameters to be estimated at each pixel position. Certainly, the problem is untraceable without any prior knowledge about unknown parameters. The second limitation of this previous work is that there is no analysis related to variation of field aberrations with depth (experimental data in this study were limited to beads mounted only on one depth).
Indeed in fluorescence microscopy, the aberrations increase as a function of depth from the coverslip \cite{Aguet_2008_PSF_modelling}, \cite[Chapter 23]{Pawley_2006_handbook}. The experimental study presented in \cite{Jezierska_phd} 
indicate that the typical for confocal microscopy intensity
decrease and effect of growing PSF size with depth are not present in macroscopy.

In this paper, we investigate the confocal macroscopy PSFs symmetry.  
We propose a procedure and experimental setup for optical center identification. Our main contribution lies in the problem formulation, its resolution and the evaluation of these results.
The experimental results show that 
our proposed model and its solution fits the experimental data well.

The paper is organized as follows. 
We present two alternative problem formulation in Section~\ref{sec:PSF_model}.
Next, in Section~\ref{sec:PSF_proposed} the related optimization methods are discussed. 
The two models are compared on synthetic data in Section~\ref{sec:PSF_results}, %
which also
illustrates 
the performance of our approaches on real data.
Finally, Section~\ref{sec:PSF_conclusions} concludes the paper.

\section{Problem statement} \label{sec:PSF_model}

\subsection{Notation} 
\begin{figure}
\begin{center}
 \scalebox{1.0} { 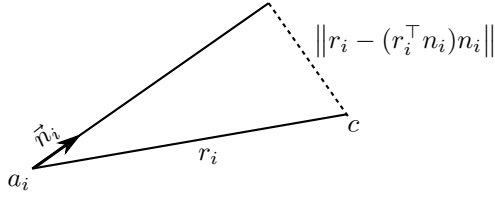}%} 
\end{center}
\caption{Distance from a point $\psfPoint$ to a line $\{a_i,n_i\}$ in arbitrary dimensions Euclidean space. \label{fig:linedistance}} 
\end{figure}

Let $\left(\psfPSF_i\right)_{i \in \left\{1, \ldots, N\right\}}$ be a finite set of identified PSFs. Let $a = (a_i)_{1 \leq i \leq N}$ where for all $i \in \left\{1, \ldots, N\right\}$ $a_i \in \RR^K$ is a center of mass of $\psfPSF_i$, and $n_i \in \RR^K$ the unit vector indicating the principal axis of inertia of $\psfPSF_i$. Since we consider 3D images, in the following $K =3$.
Let $\psfPoint \in \RR^K$ be a point. The
distance from $\psfPoint$ to the line $\{a_i,n_i\}$ is given by:
\begin{equation}
%\text{dist}(\psfPoint, \{a_i,n_i\}) = \dataTerm( a\psfPoint_i - ( a\psfPoint_i . n_i) n_i )
\text{dist}(\psfPoint, \{a_i,n_i\}) = \dataTerm( r_i - ( r_i^\top n_i) n_i )
\end{equation}
where $ r_i \in \RR^K$ is a vector from point $a_i$ to $c$ and $\dataTerm$ is some distance measure. More generally we will consider  $\dataTerm$ to be any error measure in $\Gamma_0 (\RR^K)$. Fig.~\ref{fig:linedistance} illustrates the case of $\dataTerm$ given by $\left\|\cdot\right\|$.
Thus we have
\begin{align}
r_i - ( r_i^\top n_i) n_i & = (\psfPoint-a_i) - (\psfPoint-a_i)^\top n_i n_i  \nonumber \\
                                                           & = (I- n_i n_i^\top) (\psfPoint-a_i) \nonumber\\
                                                           & = (I- n_i n_i^\top) \psfPoint- (I- n_i n_i^\top) a_i
\label{eq:distance_line_center}                                                           
\end{align}
where $I$ is the $K \times K$ identity matrix. Using the introduced notation we formulate the problem of finding the coordinates of optical center denoted by $\widehat{\psfPoint} \in \RR^K$. Note that ideally, the distance from any line 
$\{a_i,n_i\}$ to the optical center should be $0$. However, as the measurements are noisy, the problem of finding the optical center needs to be formulated in an optimization framework. 
The data include the measurements of $(a_i)_{1 \leq i \leq N}$ and $(n_i)_{1 \leq i \leq N}$.
Next, the observation model of them is presented.
 
\subsection{Observations model}
Let, for all $i \in \left\{1, \ldots, N\right\}$, $\widetilde{a}_i \in \RR^K$ and $\widetilde{n}_i \in \RR^K$ 
be vectors of observations related to an original center of mass $a_i$ and
the principal orientation $n_i$.
Next we propose two formulations which relate the observations and optical center. 

\begin{model} \label{mod:psf_1}
Let $\inputSig \in \mathcal{\inputRV} $ be a vector of unknown variables defined as $\inputSig = \psfPoint$ and  $\mathcal{\inputRV} = \RR^K$.
Let $\operator \in \mathcal{\operator} = \RR^{K \times KN}$ be defined as 
 $\operator = [\operator_1, \ldots, \operator_N]^\top$, where $\operator_i \in \RR^{K \times K}$ is given by $\operator_i = \omega_i \left(I- \widetilde{n}_i \widetilde{n}_i^\top\right)$.
For all $i \in \left\{1, \ldots, N\right\}$ $\omega_i$ denotes a positive weight.
%$m_i = n_i + \Gnoise_i$ and $\Gnoise_i$ is a zero mean Gaussian noise. 
Let $\outputSig \in \mathcal{\outputRV}$ be a vector of observations with $ \mathcal{\outputRV} = \RR^{KN}$. We define $\outputSig$ as $y = (\outputSig_i)_{1 \leq i \leq N}$ where for all $i \in \left\{1, \ldots, N\right\}$ $\outputSig_i \in \RR^K$ is a weighted vector of observations such that $\outputSig_i = \operator_i \widetilde{a}_i $.
Hence  from \eqref{eq:distance_line_center} we obtain the following model
\begin{equation}
\outputSig = (\operator+ \Delta \operator) \inputSig + \Delta \outputSig
\label{eq:PSF_model_1}
\end{equation}
where $\Delta \operator \in \mathcal{\operator}$ and $\Delta \outputSig \in \mathcal{\outputRV}$ are some error, resulting from uncertainty of measurements on $\operator$ and $\outputSig$, respectively.
\end{model} 

Introducing the variable $d_i \in \RR$, defined for all $i \in \left\{1, \ldots, N\right\}$ as
$d_i = n_i^\top (\psfPoint-a_i)$ transforms \eqref{eq:distance_line_center} into a linear functional in terms of $n_i$, $a_i$ and $\psfPoint$, i.e.
\begin{align}
r_i - (r_i^\top n_i) n_i & = \psfPoint-a_i -n_i d_i.
\label{eq:linear_distance_line_center}                                                           
\end{align}
The variable $d_i$ can be regarded as the length of the line segment connecting $a_i$ and the point given by projection of $\psfPoint$ on the line $\{a_i,n_i\}$.  
The collections of variable $(d_i)_{1 \leq i \leq N}$ forms vector $d$. Using this change of variables, we develop the following formulation.

\begin{model} \label{mod:psf_2}
Let $\outputSig \in \mathcal{\outputRV} = \RR^{KN}$ be a vector of observations 
defined as $\outputSig = ( \omega_i a_i)_{1 \leq i \leq N}$.
The related vector of unknown variables $\inputSig \in \mathcal{\inputRV} = \RR^{N+K}$
is given by $\inputSig = [\psfPoint, d]^\top$. 
Let $\operator \in \mathcal{\operator} = \RR^{NK \times (K+N)}$ be defined as $\operator = \left[\mathsf{C} \; \mathsf{D}\right]$, where $\mathsf{C} = \left(\omega_i \; I\right)_{1 \leq i \leq N}$ and $\mathsf{D} = \left(\mathsf{D}^{\left(i,j\right)}\right)_{1 \leq i \leq N, 1 \leq j \leq N}$ with non zero elements $\mathsf{D}^{\left((i-1)K+j,i\right)} = \omega_i n_i^{(j)}$. 
Using introduced notations, from \eqref{eq:linear_distance_line_center}, we obtain the following model
\begin{equation}
\outputSig = (\operator+ \Delta \operator) \inputSig + \Delta y
\label{eq:PSF_model_2}
\end{equation}
where $\Delta \operator \in \mathcal{\operator}$ and $\Delta \outputSig \in \mathcal{\outputRV}$ are some error, resulting from uncertainty of measurements on $n_i$ and $a_i$, respectively.
\end{model} 

The advantage of the first formulation over the second one is the reduced size of unknown vector $\inputSig$. Especially that usually the number of observations $N$ is much greater than the dimensionality of search space $K$, i.e. $N \gg K$. However the first formulation may lead to complex statistical properties of errors $\Delta \operator$ and $\Delta \inputSig$, which can be regarded as drawback.  
In Formulation~\ref{mod:psf_2}, an auxiliary variable $d$ is introduced, which depends on the variable $c$, whereas this dependence is not taken into account explicitly. This should lead to a suboptimal results. However, an interesting observation is that for $\dataTerm = \left\|\cdot\right\|$, we obtain as an optimal solution to the problem $\argmin_{c,d} \sum_{i=1}^N \left\| c - a_i - d_i n_i \right\|$, 
$\forall i \in \left\{1, \ldots, N\right\} \; d_i = c- a_i ^\top n_i$. Hence in such case the relation is given implicitly. For another choices of $\dataTerm$ the problem remains open. 
One method to account for outliers in both formulations is to set appropriately $\omega_i$. An interesting choice for 
$\omega_i$ could be the ratio between the first and second eigenvalue provided by PCA.

\subsection{Problem}
Next we formulate an optimization problem. The goal is to find an estimate $\widehat{\inputSig}$.
and from this solution to recover the desired estimate of the optical center $\widehat{\psfPoint}$.
 
\begin{problem} \label{prob:PSF_projection}
 Let $\left(\regTermHyb_r\right)_{1\leq r \leq R}: \RR^{P_r} \mapsto \RR $ be functions in $\Gamma_0 (\RR^{P_r})$. We want to:
\begin{equation}\label{e:PSF_problem1}
\minimize{\inputSig\in \mathcal{\inputRV}} \dataTerm \big( \operator \inputSig - \outputSig \big) + \iota_C(x)+ \sum_{r=1}^R \regTermHyb_r(\decomp_r x).
\end{equation}
where $\iota_C$ is an indicator function of $C$ and $C$ is
a closed convex subset of $\mathcal{\inputRV}$.
\end{problem}
The hereabove problem admits the following interpretation. We seek a point $\widehat{\psfPoint}$ minimizing \eqref{e:PSF_problem1}, i.e. the distance between this point and the line stemming from all measured PSFs and oriented along all the main axes of inertia, subject to some constraints. The measurement of the center of mass $a_i$ are assumed to be noisy.

\begin{problem} \label{prob:PSF_linear_formulation}

We want to:
\begin{align} 
\minimize{\substack{ \inputSig \in \mathcal{\inputRV}, \; \Delta \outputSig \in \mathcal{\outputRV}, \\ \Delta \operator  \in \mathcal{\operator}}}
& \left\| \Delta y \right\|^2 + \left\| \Delta \operator \right\|^2 \; \; \text{such that} \; \; \nonumber \\
&  \quad \quad (\operator+\Delta \operator)\inputSig - (y + \Delta y) = 0
\label{e:PSF_problem2}
\end{align}
where $\Delta b$ and $\Delta h $ model the perturbations on $b$ and $h$, respectively.
One can recover from the solution $\widehat{z}$ of the above problem, the desired estimate of the optical center $\widehat{\inputSig}$.
\end{problem}
The objective of this work is to address the Problem~\ref{prob:PSF_projection} and the Problem~\ref{prob:PSF_linear_formulation}. Note that the second problem formulation is more realistic, due to the assumption that both measurements, $a_i$ and $n_i$, are subject to noise. However, Problem~\ref{prob:PSF_projection} allows for appropriate choice of $\dataTerm$ that can handle outliers.

\section{Optimization framework} \label{sec:PSF_proposed}
In some special case, the solution of the Problem~\ref{prob:PSF_projection} admits the closed form expression. For instance, in case of Formulation~\ref{mod:psf_1}, $C = \mathcal{\inputRV}$, $R = 0$ and $ \dataTerm(u) = \sum_{i=1}^N \left\|u_i\right\|^2$ we have: 
\begin{equation}
\widehat{\inputSig} = \left(\sum_{i=1}^N  \operator_i^\top  \operator_i \right)^{-1} \left(\sum_{i=1}^N  \operator_i^\top \outputSig_i\right)
\label{eq:}
\end{equation}
So we just need to inverse $K \times K$ matrix.
More generally, the more robust choices of  $\dataTerm$ can be considered. Interesting cases could be: $\dataTerm(u) = \|u\|_1$,  $\dataTerm(u) = \|u\|$,  $\dataTerm(u) = \sum_{i=1}^N \|u_i\|$, $\dataTerm(u) = L_{t} (\|u\|)$, $\dataTerm(u) = \sum_{i=1}^N  L_{t}\left(\|u_i\|\right)$, where $L_{t}$ denotes Huber function, i.e.
 \begin{equation}
L_{t} (u) 
= 
\begin{cases}
\frac{1}{2} u^2 & \text{if} \left| u \right| \leq t \\
t \left( \left| u \right| -   0.5 t  \right) & \text{otherwise}
\end{cases}
\end{equation}
In such cases, one can resort to proximal splitting algorithms. One possibility is to use the primal-dual algorithm~\cite{Combettes_2011_primal_dual} summarized 
in Algorithm \ref{PSF_primal_dual}.
\begin{algorithm}[ht]
\caption{Primal-dual algorithm for solving~\eqref{e:PSF_problem1}. \label{PSF_primal_dual}}
\begin{algorithmic}
\STATE \textbf{Initialization:}
\STATE Set $x_0 \in \RR^K$, and $(\forall r \in \left\{0,\ldots,R\right\})$ $v_{r,0} \in \RR^{P_r}$.\\
\STATE \textbf{Iterations:} \\
\STATE For $k=0, \ldots$
\STATE
$\left \lfloor \begin{array}{l}
y_{1,k} = x_k - \gamma  \left( \operator^\top v_{0,k}+\sum_{r=1}^{R} \decomp_r^\top v_{r,k}\right) \\
p_{1,k} = \prox_{\gamma \iota_C} (y_{1,k})  \\
y_{2,0,k} = v_{0,k} + \gamma  \operator x_k\\
p_{2,0,k} = y_{2,0,k}  - \gamma\left( \prox_{\gamma ^{-1} \dataTerm }(\gamma ^{-1}y_{2,0,k} - y) + y\right)\\
q_{2,0,k} = p_{2,0,k} + \gamma  \operator p_{1,k}\\
v_{0,k+1} = v_{0,k} - y_{2,0,k} + q_{2,0,k} \\
\mbox{For} \; r=1, \ldots, R \\
\left \lfloor \begin{array}{l}
y_{2,r,k} = v_{r,k} + \gamma \decomp_{r}x_k\\
p_{2,r,k} = y_{2,r,k} - \gamma \prox_{\gamma ^{-1} \regTermHyb_r}(\gamma ^{-1}y_{2,r,k})\\
%\prox_{\gamma g_m^*}(\yb_{2,m,n} )\\
q_{2,r,k} = p_{2,r,k} + \gamma \decomp_{r} p_{1,k}\\
v_{r,k+1} = v_{r,k} - y_{2,r,k} + q_{2,r,k}
\end{array} \right.
\\
q_{1,k} = p_{1,k}-\gamma \left( \operator^\top p_{2,0,k}+\sum_{r=1}^{R} \decomp_r^\top p_{2,r,k} \right)\\
x_{k+1} = x_k - y_{1,k} +q_{1,k}
  \\
\end{array} \right.$
\end{algorithmic}
\end{algorithm}

It is worth noticing that the proximity operators of the functions of interest are given explicitly, i.e.:
\begin{align}
&\prox_{\gamma \left| \cdot \right|}(x) = 
  \begin{cases}
  \left(\left|x\right| - \gamma\right) \odot \text{sign} (x) & \text{if} \quad\left|x \right| \geq \gamma \\ 
  0 & \text{otherwise}
  \end{cases} \\
  &\prox_{\gamma \left\|  \cdot \right\|}(x) = 
  \begin{cases}
  x \left(1 - \frac{\gamma}{ \left\|x\right\|}\right) & \text{if} \quad 1 - \frac{\gamma}{ \left\|x\right\|} > 0 \\
  0 & \text{otherwise}
  \end{cases} \\
  &\prox_{\gamma L_{t}(x)} = 
  \begin{cases}
  \frac{x}{\gamma + 1} & \text{if} \quad \left|x \right| \leq  \frac{t}{\sqrt{\gamma}}  (\gamma + 1) \\ 
  x - t \sqrt{\gamma} \; \text{sign} (x)& \text{otherwise}
  \end{cases}
\end{align}
Moreover we recall from~\cite{Combettes_2010_prox_splitting_in_signal_processing} that 
proximity operator $p = \prox_{ \gamma \dataTerm } (u) $ where $\dataTerm = \sum_i \dataTerm_i (u_i)$  is equal to $p = [p_1^\top,...,p_N^\top]^\top$, where $p_i =  \prox_{ \gamma \dataTerm_i } (u_i)$.
The Problem~\ref{prob:PSF_linear_formulation} can be efficiently addressed using total least square (TLS) approach \cite{Golub_1980_TLS}, \cite{Markovsky_2007_TLS_}.

\section{Simulations} \label{sec:PSF_results}
\subsection{Synthetic data}
Here we report experimental results to the Problems~\ref{prob:PSF_projection} and ~\ref{prob:PSF_linear_formulation} using methods described in the previous section. 
The study aims at testing the performance of two approaches under the 
conditions simulating the experiment with images of 
point sources (beads) mounted on different depth. Since we consider three dimensional macroconfocal images, in all our experiments $K$ is set to $3$. 
We evaluate the performance of our approaches using $200$ randomly generated center of mass of $\psfPSF_i$ distributed at $2$ layers. For all $i \in \left\{0, \ldots, N\right\}$ coordinates $a_i^{(1)}$ and  $a_i^{(2)}$ are uniformly distributed over $\left[ 0, 2048 \right]$ while coordinate $a_i^{(3)}$ take value from set $\left\{50,250\right\}$.  
The original optical center position $\overline{c} = \left[ 1000, 1000, 5000\right]^\top$.

For all $i \in \left\{1, \ldots, N\right\}$ $\widetilde{a}_i$ and $\widetilde{n}_i $ 
are related to an original center of mass $a_i$ and
the principal orientation $n_i$
through the Bernoulli-Gaussian model of the following form
\begin{align}
%&\widetilde{n}_i = n_i + \epsilon_i u_i +(1-\epsilon_i)\Gnoise_i \quad \st \quad \left\|\widetilde{n}_i\right\| = 1,\\
&\widetilde{n}_i = n_i + \epsilon_i u_i +(1-\epsilon_i)\Gnoise_i,\\
&\widetilde{a}_i = a_i + \epsilon_i s_i + (1-\epsilon_i) t_i, 
%&\widetilde{\lambda}_i = \lambda_i + \epsilon_i o_i + (1-\epsilon_i) r_i, 
\label{eq:PSF_measurments}
\end{align}
where $\epsilon_i$ is a binary variable and
$\Gnoise_i = \left(\Gnoise_i^{(j)}\right)_{1 \leq j \leq K}$, $u_i = \left(u_i^{(j)}\right)_{1 \leq j \leq K}$, 
$s_i = \left(s_i^{(j)}\right)_{1 \leq j \leq K}$, $t_i = \left(t_i^{(j)}\right)_{1 \leq j \leq K}$ 
%$o_i = \left(o_i^{(j)}\right)_{1 \leq j \leq K}$, $r_i = \left(r_i^{(j)}\right)_{1 \leq j \leq K}$
are realizations of normally distributed random variable $\GnoiseRV_i$, $U_i$, $S_i$, $T_i$, 
 respectively, such that: 
\begin{align}
\begin{array}{ c c}
 U_i^{(j)}\sim \mathcal{N}(0,\sigma^2_1)  \quad & \quad \GnoiseRV_i^{(j)}\sim \mathcal{N}(0,\sigma^2_2) \\
S_i^{(j)}\sim \mathcal{N}(0,\sigma^2_3)  \quad & \quad T_i^{(j)}\sim \mathcal{N}(0,\sigma^2_4)
\end{array}
\label{eq:}
\end{align} 
The binary variable $\epsilon_i$ takes value according to the following rule:
\begin{equation}
\epsilon_i = 
\begin{cases}
0 & \text {if} \;  \rho_i \leq \varepsilon \\
1 & \text{otherwise}
\end{cases}
\end{equation}
$\rho_i$ is an random variable uniformly distributed in $[0,1]$ and $\varepsilon$ denotes the probability of occurrence of outliers. Thus, $\sigma_1$, $\sigma_3$ and $\sigma_2$, $\sigma_4$  denote standard deviation of inliers and outliers, respectively. 
Consequently, we have $\sigma_2 >\sigma_1$ and $\sigma_4 >\sigma_3$.
The results for $\varepsilon = 0.25$, $\sigma_1 =  0.015$,  $\sigma_2 = 0.030$,  $\sigma_3 = 30$,  $\sigma_4 = 60$ are summarized in Tables~\ref{tab:PSF_results_synthetic} and ~\ref{tab:PSF_results_synthetic_TLS}. We provide for 
all $j = \left\{1,\ldots, K\right\}$ the estimate bias and variance of $\widehat{c}$ averaged over $100$ noise realizations and normalized over true $\overline{c}^{(j)}$. The results include also mean squared error (MSE) averaged over $100$ noise realizations.  
\begin{center}
\begin{table} [ht]
\center
\small
\begin{tabular}{c  c |c| c| c| c |c|c|} \cline{3-8}
& &\multicolumn{3} {|c|} {Formulation \ref{mod:psf_1}} &  \multicolumn{3} {|c|} {Formulation \ref{mod:psf_2}}\\ \cline{3-8}
& $\dataTerm(u)$ &  $\|u\|_1 $ & $\|u\|$ &  $\L_{t} (u) $ & $\|u\|_1 $ & $\|u\|$ &  $\L_{t} (u) $ \\ \hline
\multicolumn{2} {c} {$\;$} &\multicolumn{6} {c|} {Bias ($\%$)} \\  \hline
\multicolumn{1} {c|} {\multirow{3} {*} {j}} & 1 
 &-0.03  & -0.13  & -0.02 &    0.03  & -0.03 &   0.02\\ \cline{2-8}
%& & & & & & \\ \cline{2-8}
\multicolumn{1} {c|} {}& 2 & 
 -0.30  & -0.30 &  -0.29 &    -0.18 &  -0.14 &  -0.19\\ \cline{2-8}
%& & & & &  \\ \cline{2-8}
\multicolumn{1} {c|} {} & 3 & 
 4.50 &   7.30 &    4.42 &  2.46 &    2.47 &    2.39 \\\hline
%& & & & &  \\ \hline
\multicolumn{2} {c} {$\;$} &\multicolumn{6} {c|} {Sigma ($\%$)} \\  \hline
\multicolumn{1} {c|} {\multirow{3} {*} {j}} & 1 & 
0.77  &  0.72 &   0.75  &  0.69   & 0.67 &   0.69 \\ \cline{2-8}
%& & & & & \\ \cline{2-8}
\multicolumn{1} {c|} {}& 2 & 
0.89 &    0.78 &   0.86&       0.83&    0.72&    0.81 \\ \cline{2-8}
%& & & & &  \\ \cline{2-8}
\multicolumn{1} {c|} {} & 3 & 
 4.58   & 7.37 &    4.51 &      2.58 &    2.59 &    2.51 \\ \hline \hline
\multicolumn{2} {c|} {MSE} & 
 225 & 365 &  221 &   123 &  124 & 120 \\  \hline
\end{tabular}
\caption{Optical center estimation problem results (Problem~\ref{prob:PSF_projection}) \label{tab:PSF_results_synthetic}}
\center
\small
\begin{tabular} {c c |c|c|} \cline{3-4}
& &\multicolumn{1} {|c|} {Formulation \ref{mod:psf_1}} &  \multicolumn{1} {|c|} {Formulation \ref{mod:psf_2}}\\ \hline
\multicolumn{2} {c} {$\;$} &\multicolumn{2} {c|} {Bias ($\%$)} \\  \hline
\multicolumn{1} {c|} {\multirow{3} {*} {j}} & 1 & -0.21& 0.02\\ \cline{2-4}
\multicolumn{1} {c|} {}& 2 & -0.30&  -0.06\\ \cline{2-4}
\multicolumn{1} {c|} {} & 3 &  0.53 &  -0.08 \\ \hline
\multicolumn{2} {c} {$\;$} &\multicolumn{2} {c|} {Sigma ($\%$)} \\  \hline
\multicolumn{1} {c|} {\multirow{3} {*} {j}} & 1 &0.69 &  0.66\\ \cline{2-4}
\multicolumn{1} {c|} {}& 2 &0.78 & 0.73 \\ \cline{2-4}
\multicolumn{1} {c|} {} & 3 & 0.92& 0.77 \\ \hline \hline
\multicolumn{2} {c|} {MSE} & 40&  34  \\  \hline
\end{tabular}
\caption{Optical center estimation problem results (Problem~\ref{prob:PSF_linear_formulation}) \label{tab:PSF_results_synthetic_TLS}}
\end{table}
\end{center}

One can observe that the Problem formulation given in~\ref{prob:PSF_linear_formulation} applied to Formulation~\ref{mod:psf_2} leads to the best results, i.e. the obtained results are almost unbiased and the standard deviation of estimate is lower than $1 \%$. 
Hence, we choose this approach to be confronted with a challenge originating from real data. 

\subsection{Real data}

To assess experimentally the PSF depth variation
we use the experimental described in \cite[Fig. 8.2]{Jezierska_phd}.
In the experimental sample, the beads were distributed over two layers. Images were acquired using a macro confocal laser scanning microscope (Leica TCS-LSI). 
Measurements
were done on images taken according to the following settings: pinhole
$1.0$ airy, $400$ Hz scan speed, excitation line $405/532$ nm, and emission range $534$ nm-$690$nm.

\begin{figure}
\begin{center}
 \renewcommand{\thesubfigure}{}
 \includegraphics[width=0.95\linewidth]{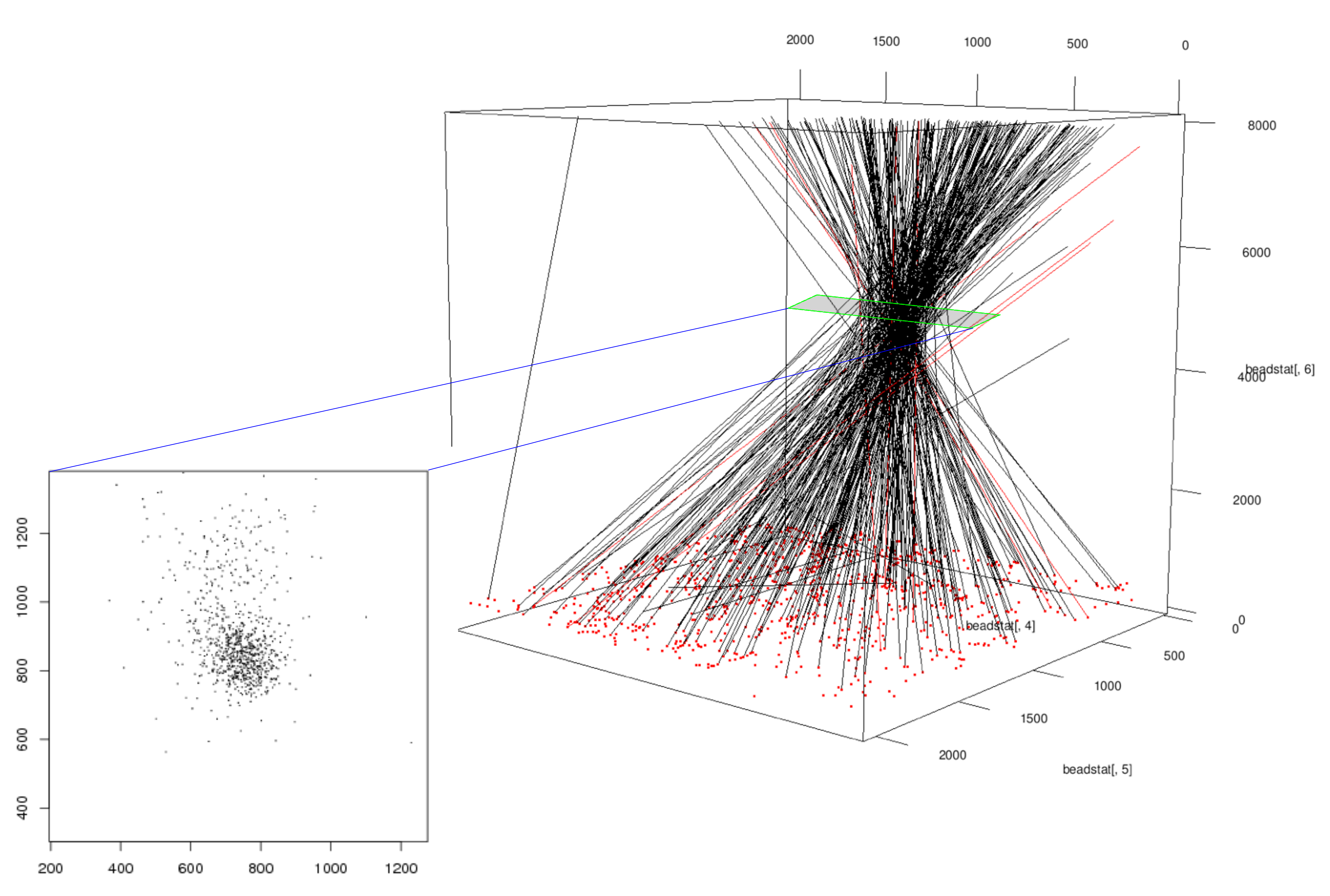}
\end{center}
\caption{ An example illustrating PSFs cone \label{fig:PSF_cone}} 
\end{figure}

As a pre-processing step, we propose to segment all the PSF in the acquired image and then to detect their center and their principal axis.
The following procedure is proposed: 
(1) Find the discrete finite set of ellipsoids in an image, which we assume to be related by injection with PSFs. (2) Compute the grey-level statistics of each independent ellipsoid. 
The difficulties that can arise in the process of identifying the signal of interest associated with PSF stem from noise, beads sticking together in the original sample or very low image SNR.
To overcome them, the following simple implementation using morphological tools is proposed.
First, the noise is reduced by anisotropic Gaussian blurring. Then 
small maxima are suppressed by volume opening~\cite{LVareaopen}, which has the effect of suppressing small objects. A top-hat operator is then used to remove low frequency variations in the background~\cite{SERRA-82}.
Next, segmentation is performed by thresholding, resulting in a binary image where $1$ correspond to the signal of interest and $0$ the background.
Finally we extend the volume of each detected nonzero ellipsoid using the Watershed algorithm~\cite{Beucher-Lantu-79}, i.e. 
we search for the maximum region around each volume under the constraint that the regions of any two PSFs may not intersect. The size of resulting volume is also limited by maximum length, width and height.
In the second step we compute the grey-level statistics of each independent ellipsoid. More specifically we use principal component analysis~\cite{Eckart_C_1936_psycom_approx_oalr} to find the center and principal axis of each ellipsoid. 
An example of the results of the above described procedure is illustrated in Fig.~\ref{fig:PSF_cone}. In the processed $12$ bit  precision image stack of size $2048 \times 2048 \times 350$ we have identified $967$ 
PSFs, within $930$ lay in the first layer and only $37$ in the second one. As expected the results indicates that the collection of lines associated with couples (the PSF center, principal axis of the PSF) form a cone like shape.
The cross-section over the PSFs cone in the the $x,y$ plane close to the optical center is illustrated in the zoomed image in Fig.~\ref{fig:PSF_cone}. Ideally, one should expect only a point in this plane.

 \begin{figure}
\centering
\includegraphics[width=0.75\textwidth]{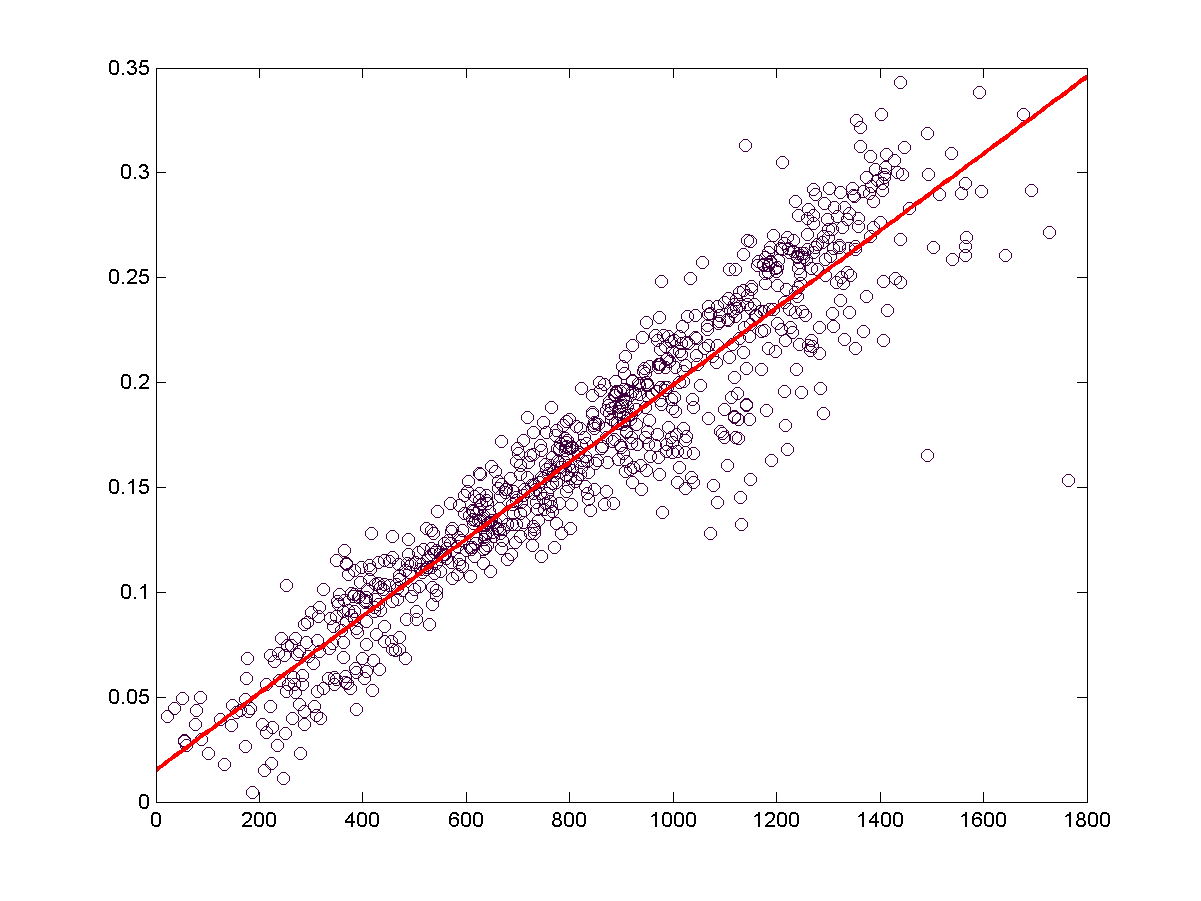}
\caption{Deviation from the vertical in radians as a function of distance to the
  optical axis.\label{fig:orient-vs-distance}}
\end{figure}

Next we identify the optical center using \ref{prob:PSF_linear_formulation} applied to Formulation~\ref{mod:psf_2}.
For $\omega_i$ set to ratio between first and second eigenvalue we obtain $\widehat{c} = [688, 887, 5201]^\top$. Fig.~\ref{fig:orient-vs-distance} illustrates  $\arccos(\widetilde{n}_i^{(3)})$ in a function of $\ell_2$ distance from $[\widetilde{a}_i^{(1)}, \widetilde{a}_i^{(2)}]^\top$ to $[\widehat{c}^{(1)}, \widehat{c}^{(2)}]^\top$ for beads whose ration between the first two eigenvalues are greater than $2.2$. The main orientation of PSF of the bead is well correlated to the distance from the bead to the optical axis. The relation is close to linear. This observation is consistent with radial symmetry of the PSFs, i.e with our main hypothesis, that all main PSF's axis converge to one point. We note that the beads furthest away from the optical axis have an orientation of nearly $0.35$ radian, i.e. almost $20^{\circ}$.

\section{Conclusions} \label{sec:PSF_conclusions}
In this paper we have presented an experimental study based on images of fluorescent bead which aim to find out if we could detect the position of the optical center towards which beads normally point. We have shown the position of the optical center to be estimated robustly in
spite of this noise and despite the presence of noticeable outliers. 
We have observed that the PSFs orientation is consistent with radial invariance.
The complexity of the problem has been reduced by applying several simplification, related to  elliptic shape of PSF and noise distribution corrupting the identified PSFs center and main directions. This assumption may be relaxed provided that more experimental data are available.

%------------------------------------------------------

\bibliographystyle{plain}
\bibliography{bibliography}

\end{document}

%% file: linedistance.pdf_tex
%% Creator: Inkscape 0.48.2, www.inkscape.org
%% Accompanies image file 'linedistance.pdf' (pdf, eps, ps)
\begingroup%
  \makeatletter%
  \providecommand\color[2][]{%
    \errmessage{(Inkscape) Color is used for the text in Inkscape, but the package 'color.sty' is not loaded}%
    \renewcommand\color[2][]{}%
  }%
  \providecommand\transparent[1]{%
    \errmessage{(Inkscape) Transparency is used (non-zero) for the text in Inkscape, but the package 'transparent.sty' is not loaded}%
    \renewcommand\transparent[1]{}%
  }%
  \providecommand\rotatebox[2]{#2}%
  \ifx\svgwidth\undefined%
    \setlength{\unitlength}{133.89711389bp}%
    \ifx\svgscale\undefined%
      \relax%
    \else%
      \setlength{\unitlength}{\unitlength * \real{\svgscale}}%
    \fi%
  \else%
    \setlength{\unitlength}{\svgwidth}%
  \fi%
  \global\let\svgwidth\undefined%
  \global\let\svgscale\undefined%
  \makeatother%
  \begin{picture}(1,0.52748569)%
    \put(0,0){\includegraphics[width=\unitlength]{linedistance.pdf}}%
    \put(-0.00342632,0.01116966){\color[rgb]{0,0,0}\rotatebox{-0.99999956}{\makebox(0,0)[lb]{\smash{$a_i$}}}}%
    \put(0.95259289,0.16351595){\color[rgb]{0,0,0}\rotatebox{-4.98786731}{\makebox(0,0)[lb]{\smash{$\psfPoint$}}}}%
    \put(0.85881246,0.39947876){\color[rgb]{0,0,0}\makebox(0,0)[lb]{\smash{$\left\| r_i - ( r_i^\top n_i) n_i\right\|$}}}%
    \put(0.09046805,0.11828789){\color[rgb]{0,0,0}\rotatebox{38.99999929}{\makebox(0,0)[lb]{\smash{$\vec{n}_i$}}}}%
    \put(0.5331065,0.08361155){\color[rgb]{0,0,0}\rotatebox{-0.99999956}{\makebox(0,0)[lb]{\smash{$r_i$}}}}%
  \end{picture}%
\endgroup%